\documentclass[10pt,letterpaper]{article}
\usepackage[top=0.85in,left=2.50in,footskip=0.75in,marginparwidth=2in]{geometry}

\usepackage[utf8]{inputenc}

\usepackage{cite}

\usepackage{nameref,hyperref}

\usepackage[right]{lineno}

\usepackage{microtype}
\DisableLigatures[f]{encoding = *, family = * }

\raggedright
\usepackage{changepage}

\usepackage[aboveskip=1pt,labelfont=bf,labelsep=period,singlelinecheck=off]{caption}

\makeatletter
\renewcommand{\@biblabel}[1]{\quad#1.}
\makeatother

\usepackage{lastpage,fancyhdr,graphicx}
\usepackage{epstopdf}
\fancyheadoffset[L]{0.75in}
\fancyfootoffset[L]{0.75in}

\usepackage{color}

\definecolor{Gray}{gray}{.25}

\usepackage{graphicx}

\usepackage{sidecap}

\usepackage{wrapfig}
\usepackage[pscoord]{eso-pic}
\usepackage[fulladjust]{marginnote}
\usepackage{amssymb}
\reversemarginpar

\begin{document}
\vspace*{0.05in}

\begin{flushleft}
{\Large
\textbf\newline{Embedding and generation of indoor climbing routes with variational autoencoder}
}
\newline
\\
Kin Ho Lo\textsuperscript{1}
\\
\bigskip
\bf{1} kin.ho.lo@cern.ch

\end{flushleft}

\section*{Abstract}
Recent increase in popularity of indoor climbing allows possible applications of deep learning algorthms to classify and generate 
climbing routes. In this work, we employ a variational autoencoder to climbing routes in a standardized training apparatus 
MoonBoard, a well-known training tool within the climbing community. By sampling the encoded latent space, 
it is observed that the algorithm can generate high quality climbing routes. 
22 generated problems are uploaded to the Moonboard app for user review. This algorithm could serve as a first step to facilitate indoor climbing 
route setting.


\section{Introduction}

Given increasing popularity of indoor climbing, we created a deep autoencoder to generate climbing routes (route setting) with a worldwide 
training apparatus Moonboard. One of the mottos of the apparatus is to set the standard for indoor training and over the years 
it has become globally recognised as one of the most effective strength training tools for rock climbing. Unlike many indoor climbing
gyms, its standardized setup and well-developed user platform provide an unique opportunity for prototyping and testing various 
supervised machine learning algorithms for route setting. Route setting is an important area in indoor climbing, and very often it challenges route settings' 
imaginations. Although fulfilling, route setting could be tedious at times. This work serves as a first step for wide applications 
of machine learning (ML) to facilitate or speed up the route setting.

Existing works for indoor climbing focus primarily on classification of route difficulties~\cite{Phillips_2012,alejandro_2017,scarff2020estimation,cheng_2019}, 
with or without using deep learning algorithms. Inspired by techniques from natural languauge processing~\cite{cer2018universal}, just as words constitute 
sentences, indoor climbing problems are composed of holds, route setting has been recently 
investigated with long short-term memory (LSTM) with sequential predictions based on previous climbing holds~\cite{andrew_2019}.
Here we adopted a different approach of embedding each climbing problem into a latent space to allow maximal freedom 
for the algorithm to learn correlations between each climbing hold. 

The paper is organized as follows. Section~\ref{sec:method} gives an overview of the algorthm. Section~\ref{sec:result} describes the results 
obtained from the training of the algorithm. 

\section{Moonboard}
\begin{figure}[ht]
	\centering
	{\includegraphics[width=0.4\textwidth]{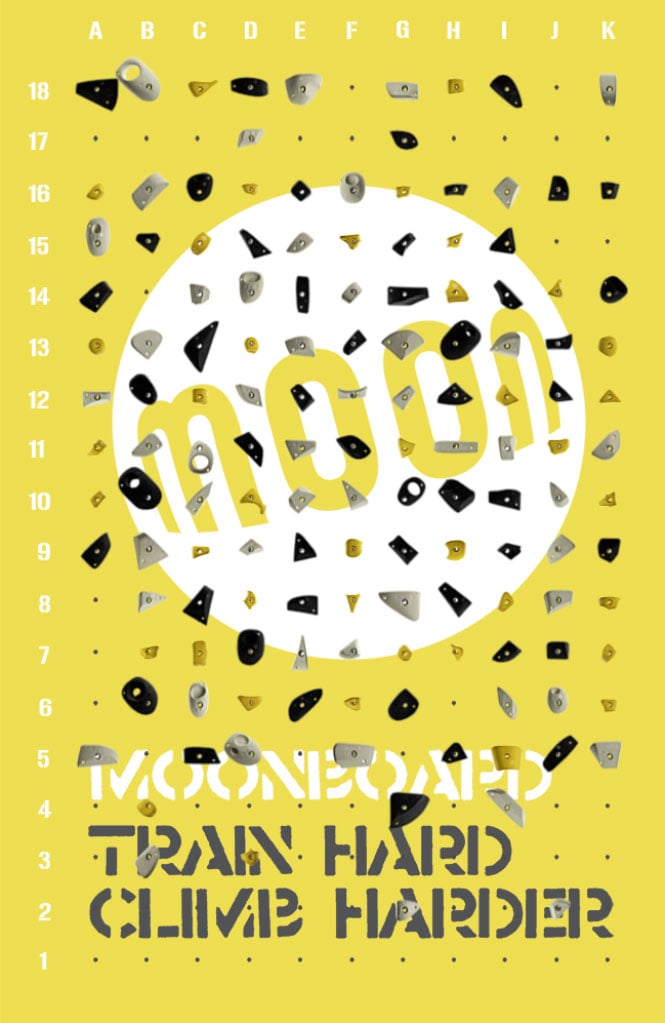}}
	{\includegraphics[width=0.4\textwidth]{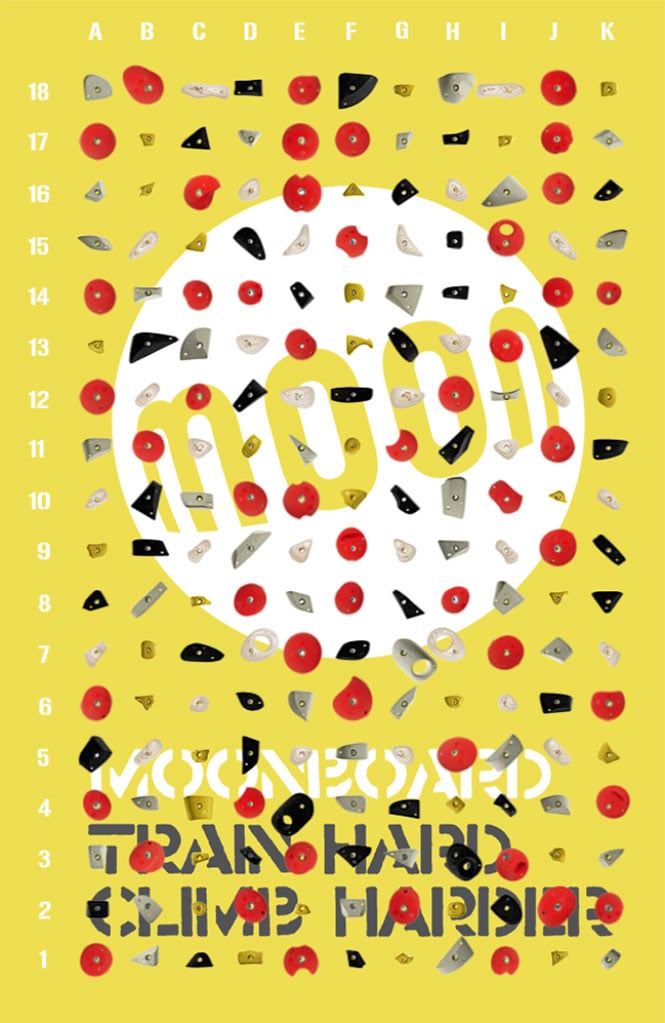}}
	\captionsetup{justification=centering}
	\caption{Moonboard setup 2016 (left), Moonboard setup 2017 (right).} 
	\label{mbsetup}
\end{figure}

The Moonboard website hosts a database of Moonboard climbing routes for each Moonboard configuration. 
Routes are represented graphically by a Moonboard image with holds selected by circles. 
Various Moonboard setups are shown in Figure~\ref{mbsetup}.
We consider the 2017 configuration, which consists of 198 holds.

\section{Method}
\label{sec:method}
Autoencoders, originally introduced in~\cite{10.5555/104279}, aims to transform inputs to outputs with least amount of deformations. More 
recently, autoencoders have been used in combination with deep learning algorithms~\cite{10.1162/neco.2006.18.7.1527,Hinton504,40d5d7fd62cb44ba934a8a75d4b2b076,10.5555/1756006.1756025}. 
We formulated our problem in terms of a variational autoencoder~\cite{kingma2013autoencoding}.

An autoencoder~\cite{scholz_2008,bengio_2007} is a mapping $f$ that has been optimized to approximate the identity on a set of
training data $x$ with the dimension of input features as $N_{d}$. It consists of a dimensional reducing mapping (encoder) $g$ followed by a 
dimension increasing mapping (decoder) $h$. Each problem is represented as 
a binary vector $x \in \mathbb{Z}_2^{N_{d}}$, $N_{d} = 198$, with each occupied hold represented as $1$ and $0$ otherwise.
\begin{equation}
    f = h \circ g \colon x \in \mathbb{Z}_2^{N_{d}} \rightarrow x_{\mathrm{out}} \in \mathbb{R}^{N_{d}}
\end{equation}
where $g$ and $h$ are defined respectively as 
\begin{equation}
    g \colon x \in \mathbb{Z}_{2}^{N_{d}} \rightarrow v \in \mathbb{R}^{N_{\ell}}
\end{equation}
\begin{equation}
    h \colon v \in \mathbb{R}^{N_{\ell}} \rightarrow x_{\mathrm{out}} \in \mathbb{R}^{N_{d}}
\end{equation}
where $N_{\ell}$ is the dimension of the latent space and $N_{\ell} = 16$ in actual training steps. 
Parameters of $g$ and $h$ are obtained by solving the optimization problem 
\begin{equation}
    \mathrm{min} \sum_{k}^{n} \left( \sum_{i}^{N_{d}} L_{\mathrm{binary}}(x_{i,k},f_{i,k}(x;\theta_{g},\theta_{h})) +
    L_{\mathrm{MSE}}(\sum_{i}^{N_{d}}x_{i,k},\sum_{i}^{N_{d}}f_{i,k}(x;\theta_{g},\theta_{h})) + 
    \sum_{i}^{N_{\ell}} L_{\mathrm{KL}}(v_{i,k})) \right) 
\end{equation}
where $n$ is the total number of input data. The first term encodes the reconstruction loss of the algorithm. 
The second term constrains the number of occupied holds in the problem and the third term is the Kullback–Leibler divergence.

\section{Result}
\label{sec:result}
In this work, we adopted the 2017 Moonboard setup with 198 holds. The dimension of the latent space is 16.
The variational autoencoder is trained with a single GPU (GeForce RTX 2080) for 2000 epochs with batch size of 512, using a training sample with 16979 Moonboard 
problems and a testing sample with 1886 problems. 

\begin{figure}[ht]
	\centering
	{\includegraphics[width=0.3\textwidth]{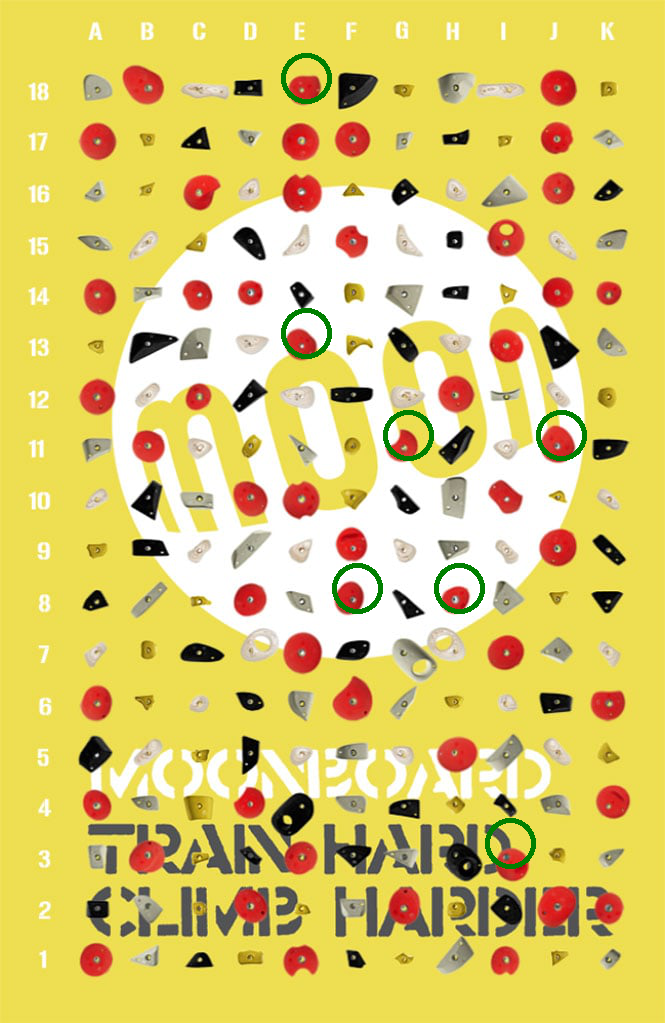}}
	{\includegraphics[width=0.3\textwidth]{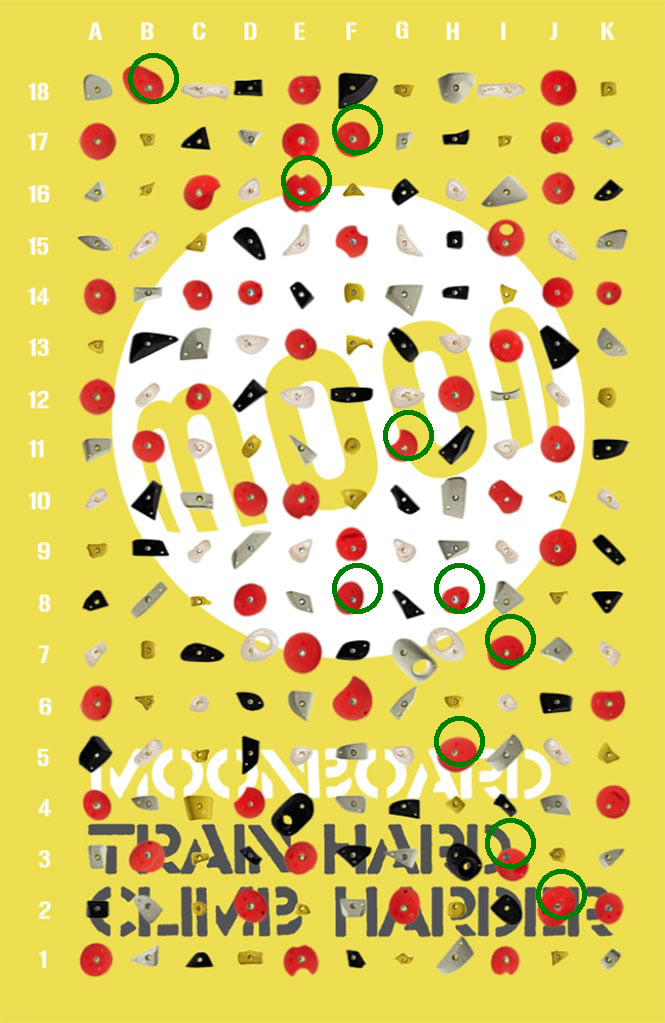}}
	{\includegraphics[width=0.3\textwidth]{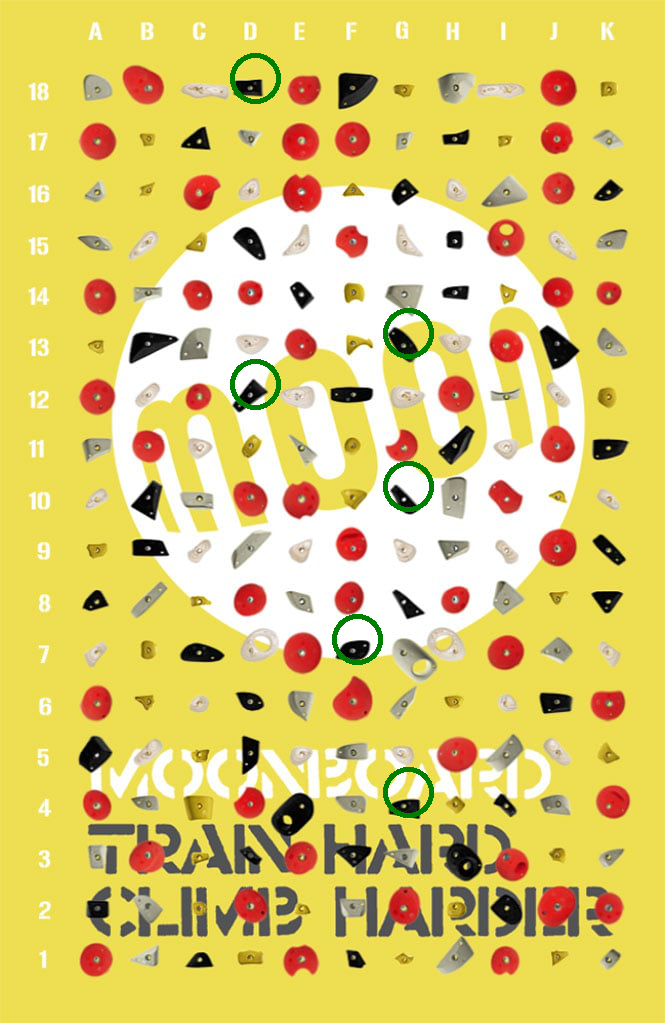}}
	{\includegraphics[width=0.3\textwidth]{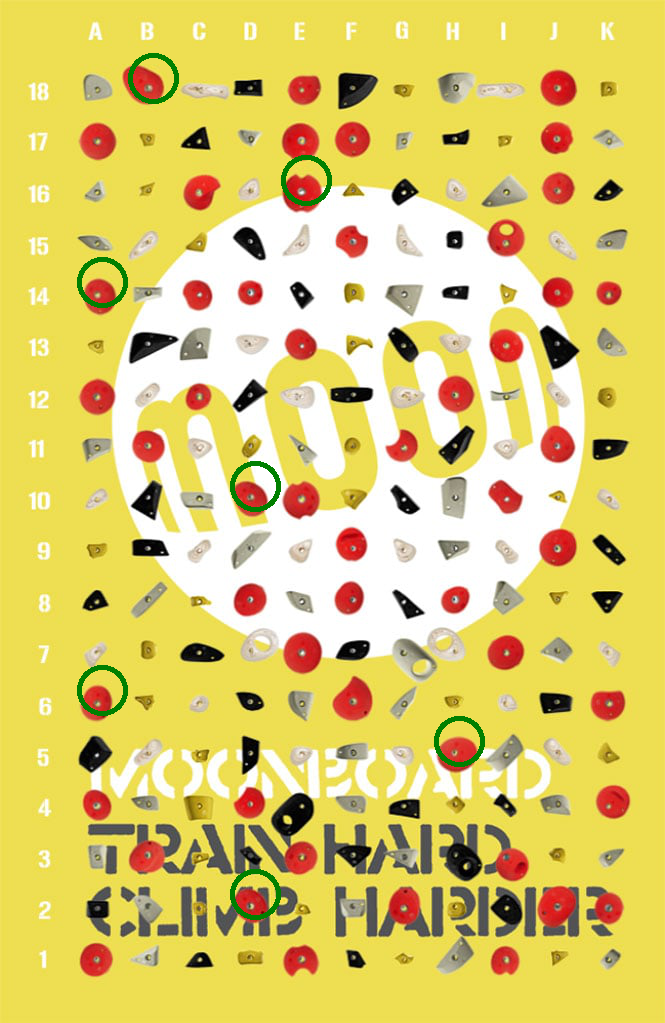}}
	{\includegraphics[width=0.3\textwidth]{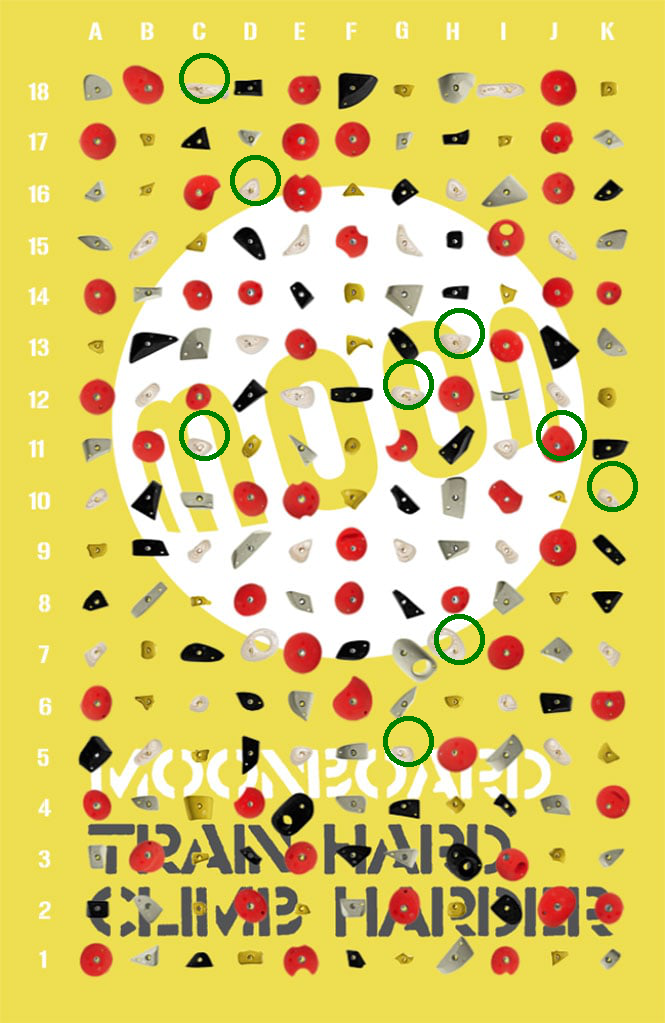}}
	{\includegraphics[width=0.3\textwidth]{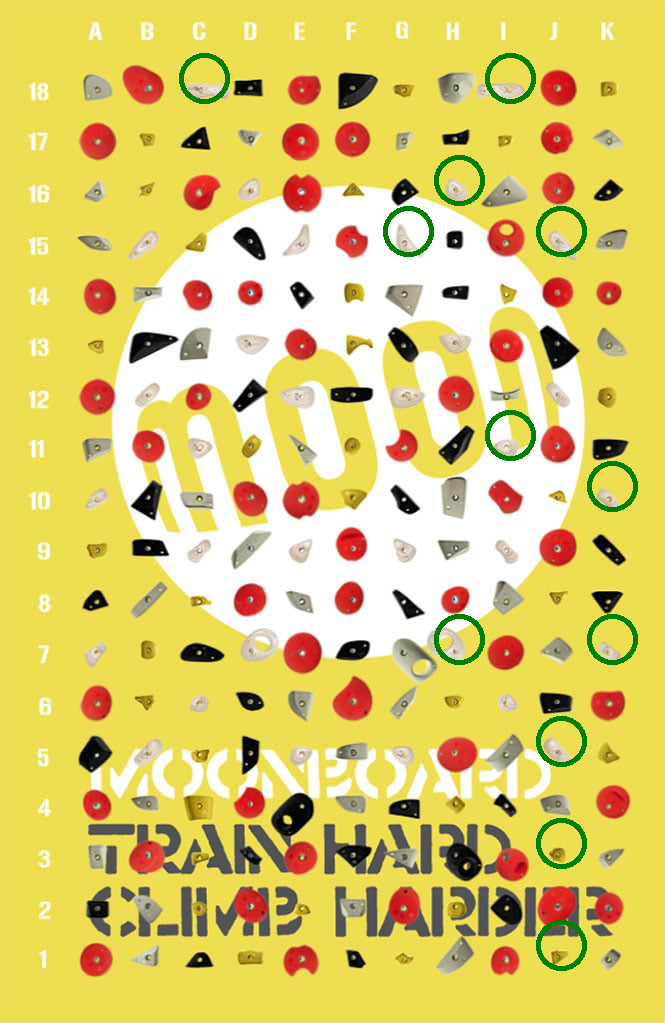}}
	{\includegraphics[width=0.3\textwidth]{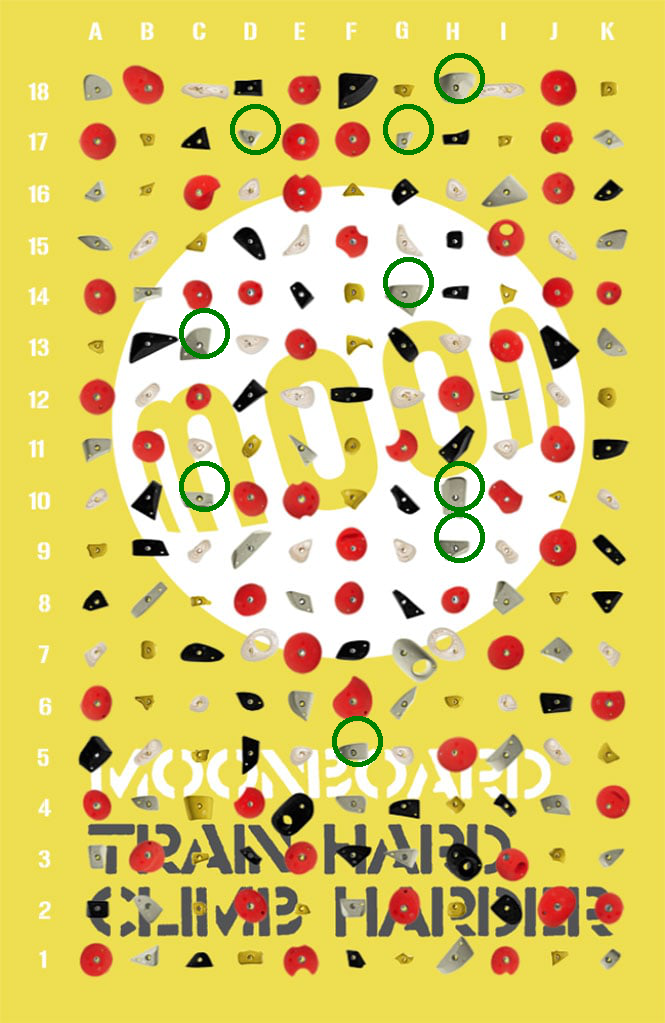}}
	{\includegraphics[width=0.3\textwidth]{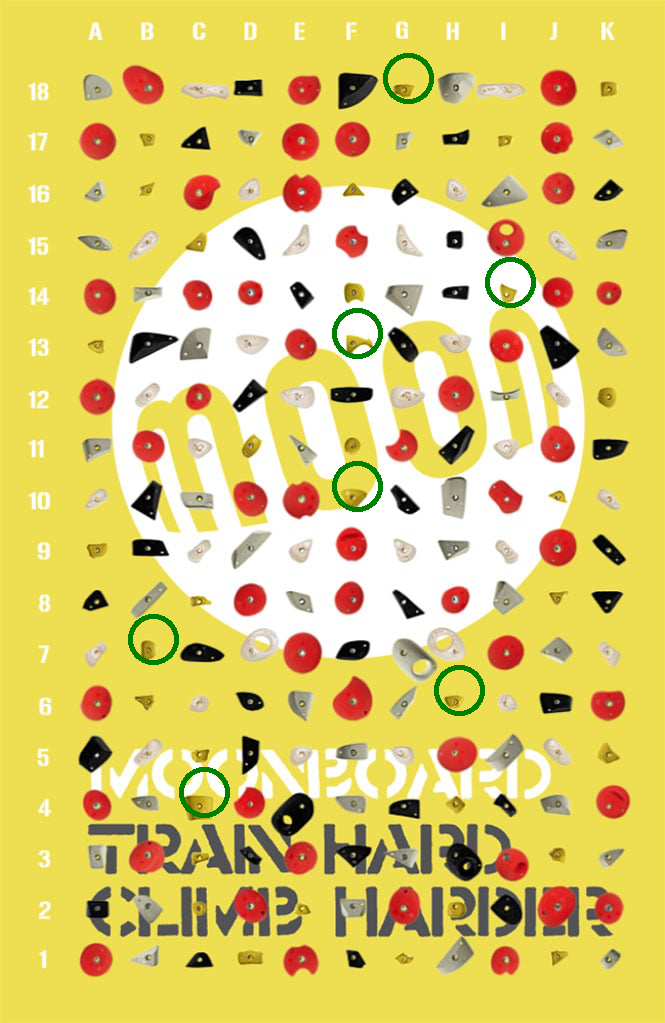}}
	{\includegraphics[width=0.3\textwidth]{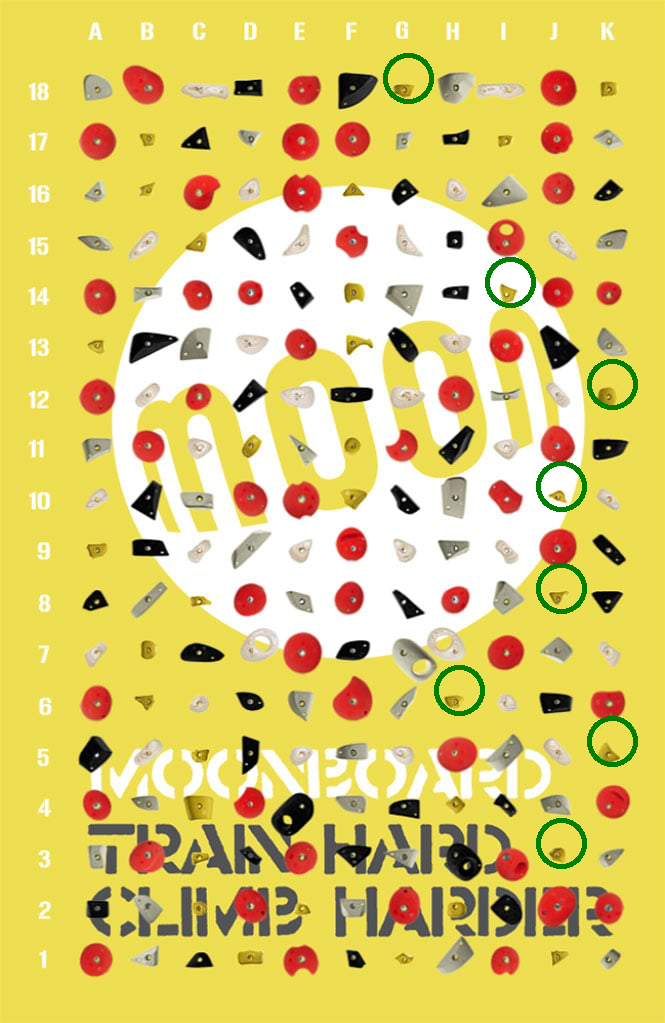}}
	\captionsetup{justification=centering}
	\caption{Examples of generated problems using only certain type of holds in the board.} 
	\label{problem1}
\end{figure}

\begin{figure}[ht]
	\centering
	{\includegraphics[width=0.3\textwidth]{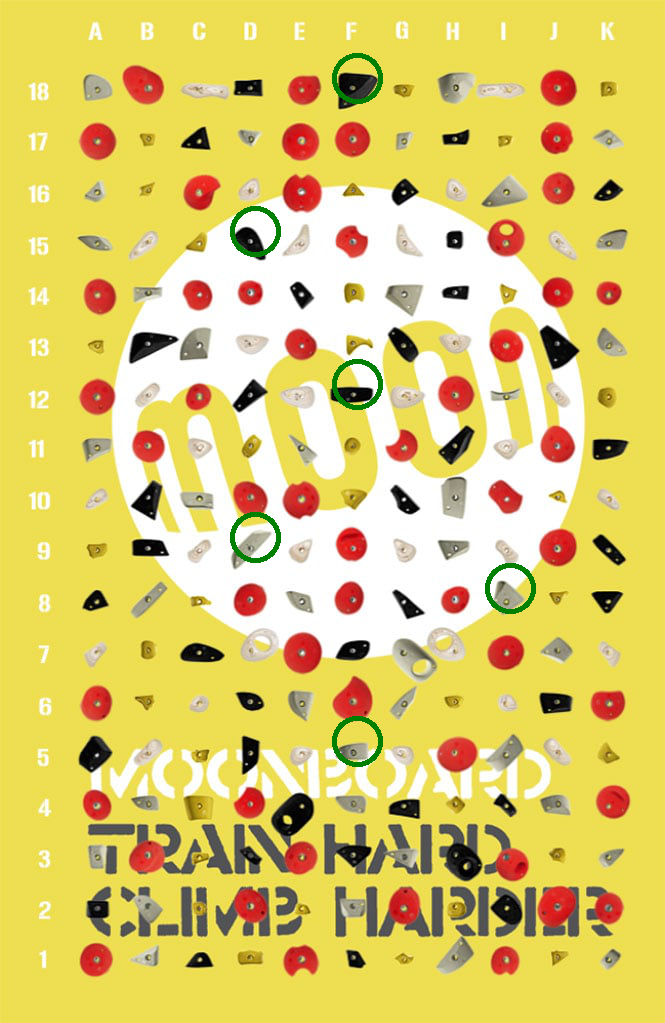}}
	{\includegraphics[width=0.3\textwidth]{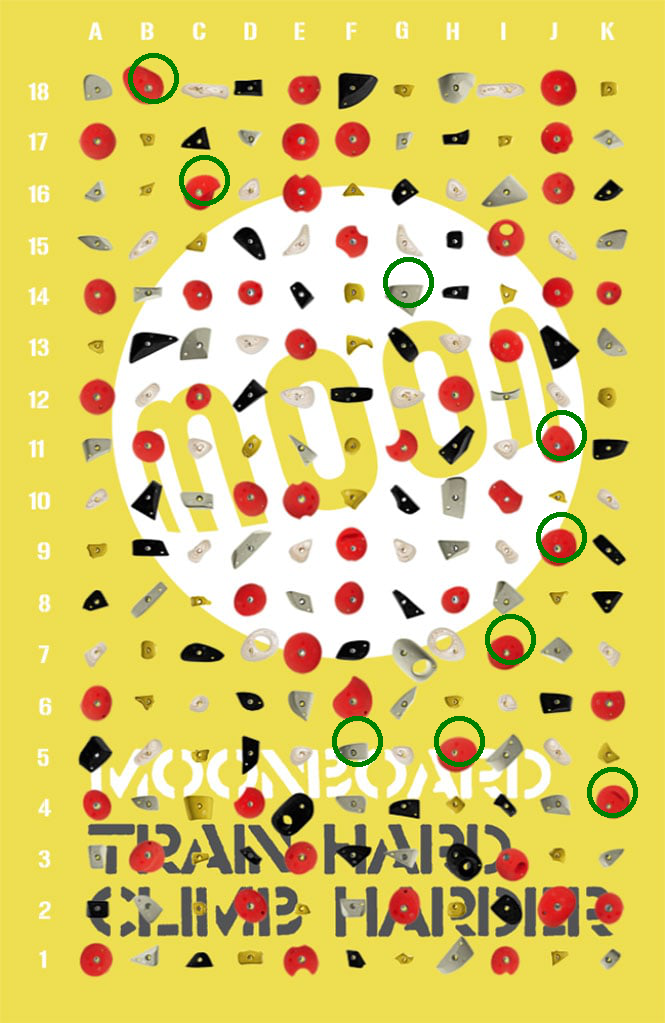}}
	{\includegraphics[width=0.3\textwidth]{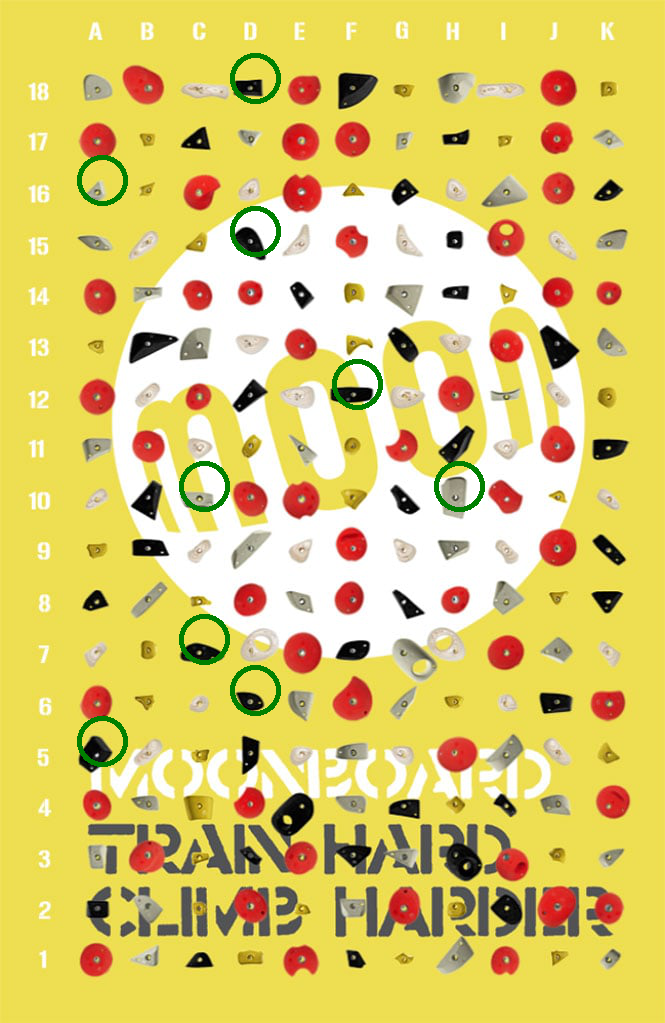}}
	\captionsetup{justification=centering}
	\caption{Examples of generated problems using relatively good holds in the board.} 
	\label{problem3}
\end{figure}

Each sample is formed by randomly selecting the existing Moonboard problems to ensure that problems with various style 
and difficulty could be considered in the algorithm. After training, 50 routes are generated by randomly sampling the latent space and selecting the holds with the 
highest probabilities according to the length indicated. All problems generated are required to have at least 6 holds. We observed that this requirement selects 27 out of 50 problems. 
22 moonboard problems are generated and uploaded to the moonboard mobile application for user review. Out of the remaining 5 problems, 1 problem coincides with an existing problem in the Moonboard database and 
4 problems are failed attempts, as described below.
Some interesting features could be observed in the generated problems suggesting that the algorithm have picked up special features of the board and the climbing routes: 
\begin{itemize}
    \item Every Moonboard climbing route has one or two finishing holds at the top row of the board. Most of generated problems exhibit this feature.
    \item Some generated problems used only a specific type of holds in the board, as shown in Figure~\ref{problem1}.
    \item There are much more problems in lower grade of difficulty. Therefore, many generated problems have relatively good holds in the board, as shown in Figure~\ref{problem3}.
\end{itemize}

\begin{figure}[ht]
	\centering
	{\includegraphics[width=0.3\textwidth]{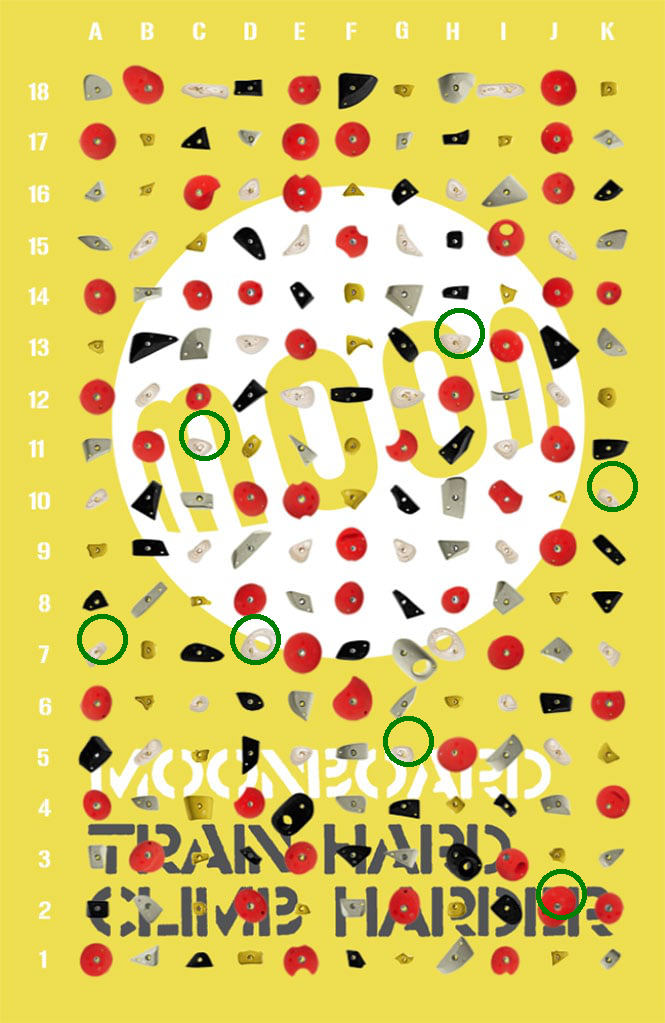}}
	{\includegraphics[width=0.3\textwidth]{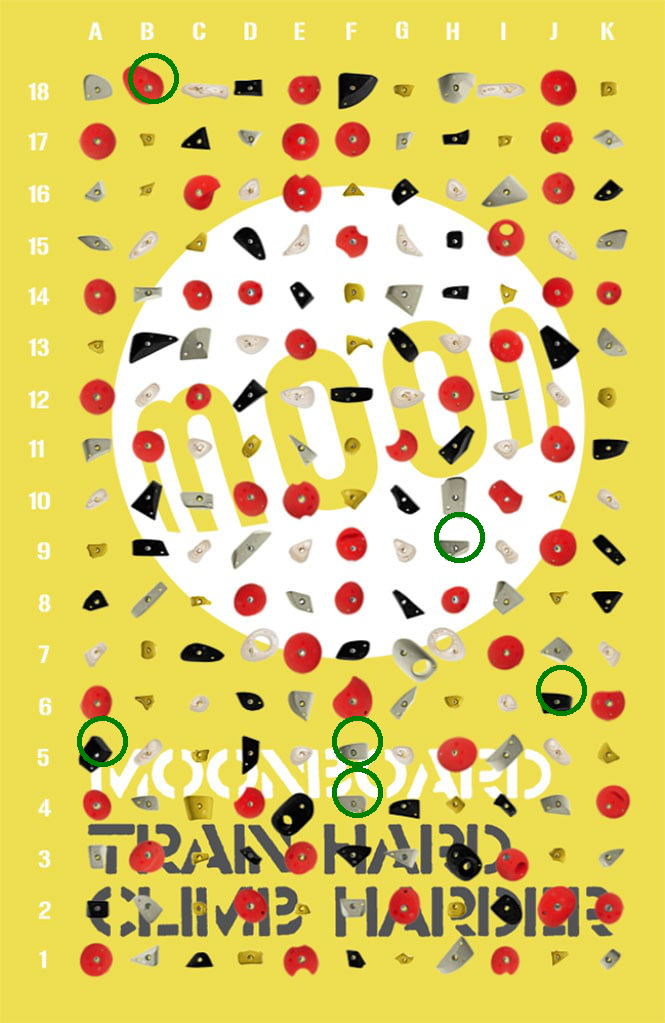}}
	{\includegraphics[width=0.3\textwidth]{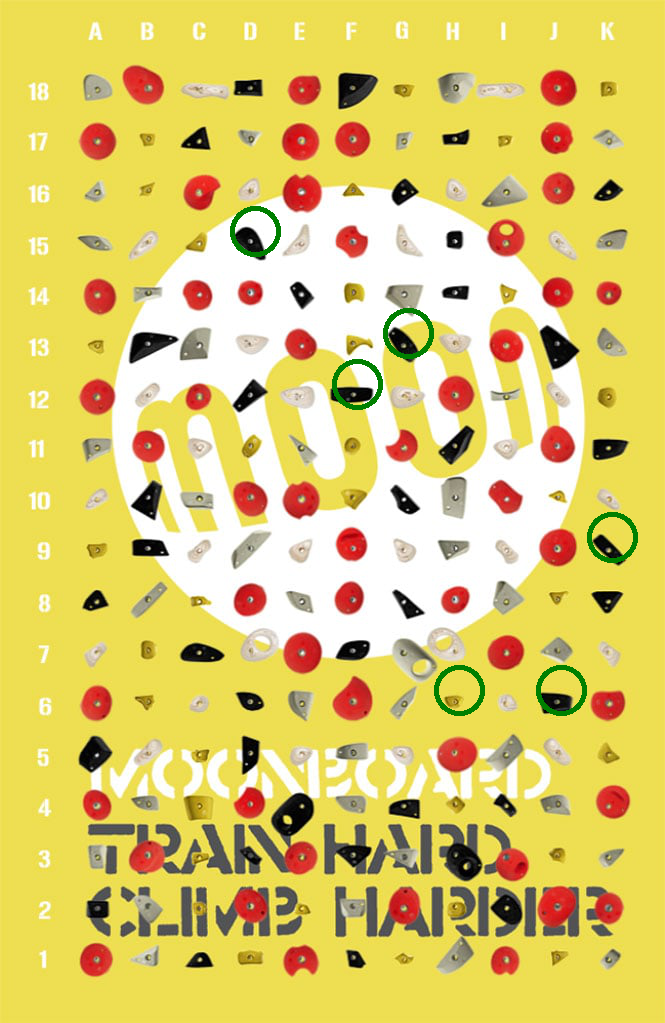}}
	{\includegraphics[width=0.3\textwidth]{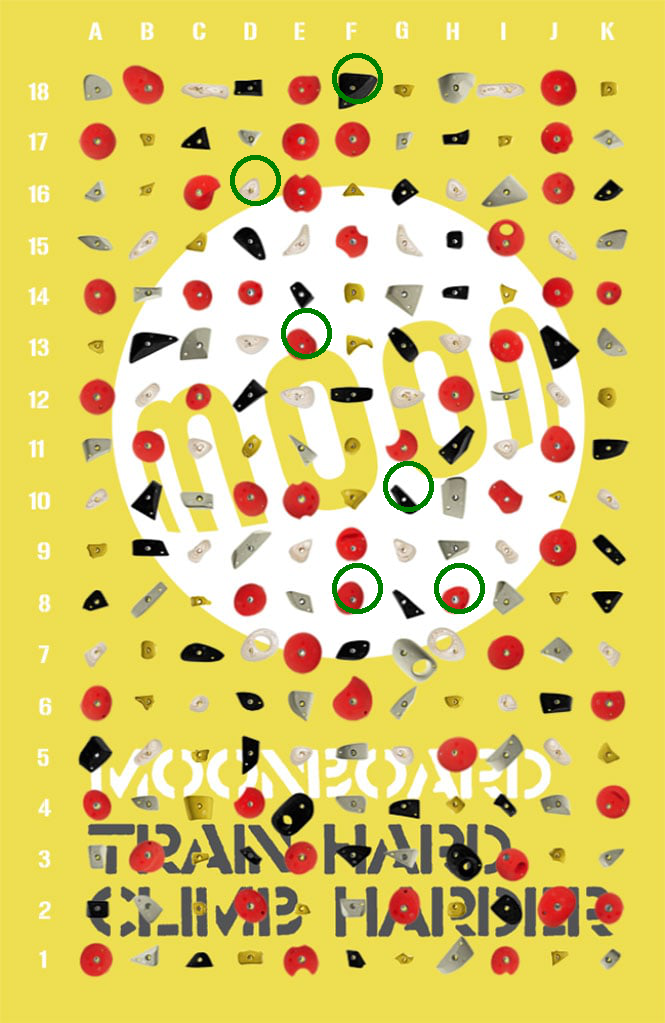}}
	\captionsetup{justification=centering}
	\caption{Examples of generated problems with bad quality.} 
	\label{problem4}
\end{figure}

Despite the fact that many problems are generated with good quality, 4 generated problems, shown in Figure~\ref{problem4} 
remain "unclimbable". These observations are highlighted below:
\begin{itemize}
    \item 1 problem is generated such that the finishing hold is almost unreachable from the rest of the holds.
    \item 2 problem are generated with missing finishing holds.
    \item 1 problem is generated with missing starting holds. Starting holds are required to be below row 7 on the board. Despite this caveat, the problem can still be climbed.
\end{itemize}

\section{Conclusions and future work}
In this work, we presented a simple application of variational autoencoder to the generation of indoor climbing routes. We found that 
the simple and well-established algorithm achieved reasonable performance and can generate climbing routes with high quality. 
Given more time, this approach could be applied to the problems with other board setup, e.g. Moonboard 2016 and Kilterboard. In the future, 
various improvements could be incorporated by, but not limited to, conditioning the algorithm with the user-specified route difficulty, climbing 
style, and enhancing the generation efficiency.

\nolinenumbers

\bibliography{paper}
\bibliographystyle{abbrv}

\end{document}